\documentclass[preprint,12pt]{elsarticle}
\usepackage{times}
\usepackage{epsfig}
\usepackage{graphicx}
\usepackage{amsmath}
\usepackage{amssymb}
\usepackage[noend]{algpseudocode}
\usepackage[ruled,vlined,linesnumbered]{algorithm2e}
\usepackage{subfig}
\usepackage{setspace}
\usepackage[pagebackref=true,breaklinks=true,letterpaper=true,colorlinks,bookmarks=true]{hyperref}

\usepackage{amssymb}

\journal{Neurocomputing}

\begin{document}

\begin{frontmatter}



\title{SD-GAN: Structural and Denoising GAN reveals facial parts under occlusion}

\author{Samik Banerjee}
\ead{samik.banerjee.howrah@gmail.com}
\author{Sukhendu Das}
\ead{sdas@iitm.ac.in}


\address{Dept. of CS\&E, IIT Madras, Chennai, India}

\begin{abstract}
Certain facial parts are salient (unique) in appearance, which substantially contribute to the holistic recognition of a subject. Occlusion of these salient parts deteriorates the performance of face recognition algorithms. In this paper, we propose a generative model to reconstruct the missing parts of the face which are under occlusion. The proposed generative model (SD-GAN) reconstructs a face preserving the illumination variation and identity of the face. A novel adversarial training algorithm has been designed for a bimodal mutually exclusive Generative Adversarial Network (GAN) model, for faster convergence. A novel adversarial "structural" loss function is also proposed, comprising of two components: a holistic and a local loss, characterized by SSIM and patch-wise MSE. Ablation studies on real and synthetically occluded face datasets reveal that our proposed technique outperforms the competing methods by a considerable margin, even for boosting the performance of Face Recognition.

\end{abstract}


\end{frontmatter}


\section{Introduction}
\label{sec:intro}
Faces appearing under occlusion is a major hindrance for accurate Face Recognition (FR), which has been far from being solved. With the advent of generative adversarial models \cite{goodfellow2014generative} in the field of deep learning (DL), there has been a surge of techniques to predict the missing values or pixels in an image. Revealing of missing parts of an image is a common image editing operation, which aims to fill the missing or masked regions in images with appropriate contents that appears to be visually realistic. The generated contents can either be as accurate as the original, or simply fit well within the context such that the restored image looks perceptually plausible and complete. Recent image completion techniques \cite{barnes2009patchmatch,huang2014image} rely on low and mid-level cues for the generation of the missing patches in the image.

Contrary to the recent techniques, our proposed method reconstructs a full face despite the fact that certain salient and unique features on the faces are occluded. The processes concerned with generating missing patches on the faces make an assumption that the similar patterns do not exist everywhere. Inline with his assumptions, Generative Adversarial Networks (GANs) aim to perform well in generating the facial parts behind the mask, due to its capability of generating the unseen. Wright \textit{et al.} \cite{wright2009robust} used a method for sparse recovery of signals for image completion, which is further used in face completion. Recently, Ren \textit{et al.} \cite{ren2015shepard} used Convolutional neural networks (CNN) for inpainting of images. Li \textit{et al.} \cite{li2017generative} used a generative model to restore face-parts occluded by patches on the  CelebA dataset, but they did not provide any result on real-world occluded face datasets, like the AR face database \cite{martinez1998ar}. They also relied on post-processing of the images to produce semantically correct images. The Generative Face Completion (GFC) \cite{li2017generative} process requires significantly large amount of training time to reach the equilibrium point.

With the aim of designing an end-to-end framework for generating face images from the masked ones, the primary contribution of this paper lies in design of a novel bimodal training algorithm for GAN. Mode-I of the training process produces faces with ambient illumination, while Mode-II denoises that generated by Mode-I. A unique training algorithm is proposed with faster convergence. An adversarial "structural" loss is also proposed in this paper in order to maintain the holistic quality of the face images. This "structural" loss consists of two components: "Structural Similarity (SSIM) loss" and "Patch-wise Mean squared error (PMSE)". The SSIM \cite{wang2004image} takes care of the holistic features of the face, while PMSE takes care of the pixel-wise differences in the faces. Further, our model converges to an equilibrium in Mode-II faster than other generative models \cite{krizhevsky2012imagenet}, since the generator is based on a denoising auto-encoder \cite{vincent2008extracting} model. The generated faces boost the performance of FR on occluded faces, when compared with the works published recently in literature.

Sections \ref{sec:GAN} and \ref{sec:DAE} give brief overviews of GAN and Denoising Auto-encoder, respectively, while section \ref{sec:loss} discusses the loss functions used in this paper. Section \ref{sec:arch} gives the details of the proposed architecture of SD-GAN, followed by the description of the proposed training algorithm in section \ref{sec:train}. In section \ref{sec:res}, the quantitative and qualitative results of our experiments, showing the effectiveness of our proposed method are reported, along with the different benchmark datasets used for experimentations. Finally, the paper concludes in section \ref{sec:conc}.

\section{Generative Adversarial Networks (GAN)}
\label{sec:GAN}
Generative Adversarial Network (GAN) \cite{goodfellow2014generative} consists of two models: the generative ($G$) and the discriminator ($D$). The CNN based deep network in $G$ captures the true data distribution, $p_{data}$, and generates images sampled from a distribution $p_{z}$, the distribution of the training data provided as input to $G$. $D$ as a counter-part of $G$ (also CNN-based) discriminates between the original images, sampled from $p_{data}$, and the images generated by $G$. Typically, $G$ learns to map from a latent space ($p_z$) to a particular data distribution ($p_{data}$) of interest, while $D$ discriminates between instances from $p_{data}$ and candidates produced by the generator. The objective of training $G$ is to increase the error rate of $D$ (\textit{i.e.}, "fool" $D$ by producing novel synthesized instances that appear to have come from $p_{data}$). This adversarial training adopted for GAN is derived from that in Schmidhuber \cite{urgen1992learning}. In other words, an alternate training procedure is performed on GAN, where $D$ and $G$ play a two-player minimax gaming strategy of a zero-sum game with the value function $V(G,D)$. The overall objective function minimized by GANs \cite{goodfellow2014generative}, is given as:
\begin{equation}
\begin{split}
\min_G \max_D V(G,D) & = \mathbb{E}_{x\sim p_{data}}[\log D(x)]\\
& + \mathbb{E}_{x\sim p_z}[\log(1-D(G(z)))]
\end{split}
\end{equation}
To learn $p_z$ over data $x$, a mapping to data space is represented as $G(z; \theta_g)$, where $G$ is a differentiable function representing a CNN with parameters $\theta_g$. Another CNN based deep network represented by $D(x; \theta_d)$ outputs a single scalar $[0/1]$. $D(x)$ represents the probability that $x$ came from the true data rather than $p_z$. 

Two major drawbacks of an adversarial system are:
\begin{enumerate}
	\item GANs can generate all the pixels in one shot, rather than guessing the value of one pixel given another pixel. This is the main reason for the noise in the output images, whenever missing pixels are generated.
	\item Reaching the Nash equilibrium \cite{nash1950equilibrium} of a game requires large number of iterations/epochs due to the instability inherent in GANs \cite{goodfellow2014generative}.
\end{enumerate}
An aim to overcome the above two drawbacks, forms the basic motivation of our work presented in this paper. To deal with noise, a Denoising Auto-encoder based generator model has been introduced in conjunction with the standard GAN framework. Further, the Mode-II reaches the Nash equilibrium faster than Mode-I. A trade-off has been done at Mode-I between the structural loss and training time, where the generator loss is thresholded for the generated images passed to Mode-II for denoising.
\section{Denoising Auto-encoder}
\label{sec:DAE}
The general deep auto-encoder, as proposed by Bengio \textit{et al.} \cite{bengio2007greedy}, maps an input vector $\vec{x} \in [0,1]^d$ to a latent representation $\vec{y} \in [0,1]^{d'}$ through a deterministic mapping $\vec{y} = f_\theta(x) = s(\textbf{W}\vec{x}+\vec{b})$ with $\theta = \{\textbf{W},\vec{b}\}$, and then maps back to the reconstructed vector, $\vec{z} = g_{\theta'}(y) = s(\textbf{W}'\vec{y}+\vec{b}'), \vec{z} \in [0,1]^d$ in the input space with $\theta' = \{\textbf{W}',\vec{b}'\}$, where $s(\cdot)$ denotes the activation function. The optimization of the parameters is based on the mean reconstruction error \cite{bengio2007greedy}:
\begin{equation}
\begin{split}
\theta^*,\theta'^* & = arg\min_{\theta,\theta'}\frac{1}{n}\sum_{i=1}^{n}L\big(\vec{x}^{(i)}, \vec{z}^{(i)}\big)\\
&=arg\min_{\theta,\theta'}\frac{1}{n}\sum_{i=1}^{n}L\big(\vec{x}^{(i)}, g_{\theta'}(f_\theta(\vec{x}^{(i)}))\big)
\end{split}
\label{eq:ae}
\end{equation} 
where, $\vec{x}^{(i)}$ represents the $i^{th}$ training sample and $L$ is the squared error $L(\vec{x},\vec{z}) = \|\vec{x}-\vec{z}\|^2$.

Vincent \textit{et al.} \cite{vincent2008extracting} designed a denoising autoencoder by modifying the formulation in equation \ref{eq:ae}. The authors assumed $\vec{\tilde{x}}$ to be a noisy approximation of $\vec{x}$, characterized by a stochastic mapping $\vec{x}\sim q_D(\vec{\tilde{x}}|\vec{x})$. The joint distribution is given as $q^0(\vec{x}, \vec{\tilde{x}}, \vec{y}) = q^0(\vec{x})q_D(\vec{\tilde{x}}|\vec{x}) \delta_{f_\theta(\vec{\tilde{x}})}(\vec{y})$, where $\delta_u(v)=0$, when $u\ne v$, and parameterized by $\theta$. Thus, $\vec{y}$ becomes the deterministic function of $\vec{\tilde{x}}$. The objective function in equation \ref{eq:ae} thus transforms into:
\begin{equation}
arg\min_{\theta,\theta'} \mathbb{E}_{q^0(\vec{x}, \vec{\tilde{x}})}\big[L\big(\vec{x}^{(i)}, g_{\theta'}(f_\theta(\vec{\tilde{x}}^{(i)}))\big)\big]
\end{equation}
Patch-wise minimization of mean-squared error (discussed later in section \ref{sec:psme}) further helps in image denoising \cite{lee2012mmse}. Thus patch-wise mean squared error loss has been used in this paper as a component of the loss function in both the generators ($G_1$ \& $G_2$) of our SD-GAN framework.


%
%
\section{Loss Functions}
\label{sec:loss}
The process of training the SD-GAN consists of two modes, and optimizes four adversarial loss functions described (later) in equations \ref{eq:d1_loss}-\ref{eq:g2_loss}. The corresponding criteria are described in the following sub-sections.

\subsection{Binary Cross Entropy Loss}
\label{sec:bce}
Binary cross-entropy is a loss function used effectively in the field of deep learning for binary classification problems and sigmoid output units. The binary class labels used at the discriminators are $0$ \& $1$, representing the real and fake (generated) images. The loss function is given as:
\begin{equation}
\begin{split}
\mathcal{L}_{bce}(\vec{\tilde{y}},\vec{y}) & = -\frac{1}{n}\sum_{i=1}^n \left[y_i \log(\tilde{y}_i) + (1-y_i) \log(1-\tilde{y}_i)\right]\\
&= -\frac{1}{n}\sum_{i=1}^n\sum_{j=1}^m y_{ij} \log(\tilde{y}_{ij})
\end{split}
\end{equation}
where, $i$ indexes $n$ samples/observations and $j$ indexes $m$ classes, and $y_i$ is the sample label (binary for LHS, one-hot vector on the RHS) and the prediction of sample is $\tilde{y}_{ij}\in(0,1):\sum_{j} \tilde{y}_{ij} =1, \text{ } \forall i,j$.

\subsection{SSIM Loss}
\label{sec:ssim_loss}
SSIM \cite{wang2004image} gives the structural similarity index between two images ($x_1$ and $x_2$). We first define SSIM index \cite{wang2004image}, estimated using multiple patches (windows) of an image. This measure between two windows $p$ and $q$ of common size $N \times N$ is:
\begin{equation}
SSIM(p,q) = \frac{(2\mu_p\mu_q + c_1)(2\sigma_{pq} + c_2)}{(\mu_p^2 + \mu_q^2 + c_1)(\sigma_p^2 + \sigma_q^2 + c_2)}
\label{eq:ssim}
\end{equation}
where, $\mu_p, \mu_q$ are the pixel-wise averages of image patches $p$ and $q$ respectively, $\sigma_p^2, \sigma_q^2$ their respective variances, $\sigma_{pq}$ the covariance of $p$ and $q$; $c_1 = (k_1L)^2$, $c_2 = (k_2L)^2$ as two variables used to stabilize the division with weak denominator, $L$ the dynamic range of the pixel-values (typically this is $2^{\#bits/pixel}-1$), and $k_1 = 0.01$ and $k_2 = 0.03$ set by default. The SSIM loss ($\mathcal{L}_{ssim}$) function estimated between two single-channel (gray-scale) images, produces a maximum value of $1$ for two identical images and decreases henceforth as the similarity between the images decreases. Hence, the SSIM loss is calculated as: 
\begin{equation}
\mathcal{L}_{ssim} = 1-SSIM(x_1, x_2)
\end{equation}
where, SSIM is given in equation \ref{eq:ssim}. Minimization of this loss provides a better estimate of the $x_2$ for $x_1$.

\subsection{Patch-wise MSE Loss}
\label{sec:psme}
Patch-wise MSE (PMSE) loss is derived as the mean-squared error between two images. Let $h_1$ and $h_2$ be the two patches extracted from $x_1$ and $x_2$, respectively. The PMSE between $x_1$ and $x_2$, is calculated as:
\begin{equation}
\mathcal{L}_{pmse}(x_1,x_2) = \sum_{i=1}^{|C|} \frac{\lambda_i}{|h|}\sum_{j=1}^{|h|} \|h_1^{(i,j)}-h_2^{(i,j)}\|^2
\end{equation}
where, $|C|$ \& $|h|$ are the number of channels and patches in an image, while $h_k$ is a patch extracted from $x_k$ and $\lambda_i$'s  are the channel-wise weights of the image ($\lambda = \{0.2989,0.5870,0.1141\}$ as given in \cite{johnson2006stephen}). A weighted linear combination (using $\lambda$) of the MSE's is used to estimate the MSE of each patch. PMSE is the average MSE over all the pair of corresponding (spatially) patches in the images.

\subsection{Structural Loss}
\label{sec:struct}

This paper also proposes a novel structural loss ($\mathcal{L}_{st}$) in addition to the binary cross-entropy loss as in DCGAN \cite{goodfellow2014generative}. The primary aim of proposing this novel loss is to constrain the structure of the generated image. The SSIM (see section \ref{sec:ssim_loss}) loss accounts for the facial structure while a mean-squared error (MSE) based loss applied patch-wise (refer section  \ref{sec:psme}) helps to replicate of the illumination variation in $G_1$ and denoising in the auto-encoder based $G_2$. The structural loss is given as:
\begin{equation}
\mathcal{L}_{st} = \frac{\mathcal{L}_{ssim} + \mathcal{L}_{pmse}}{2}
\label{eq:st}
\end{equation}


\section{The proposed architecture: SD-GAN}
\label{sec:arch}

\begin{figure*}[!htbp]
	\centering
	\includegraphics[width = \linewidth]{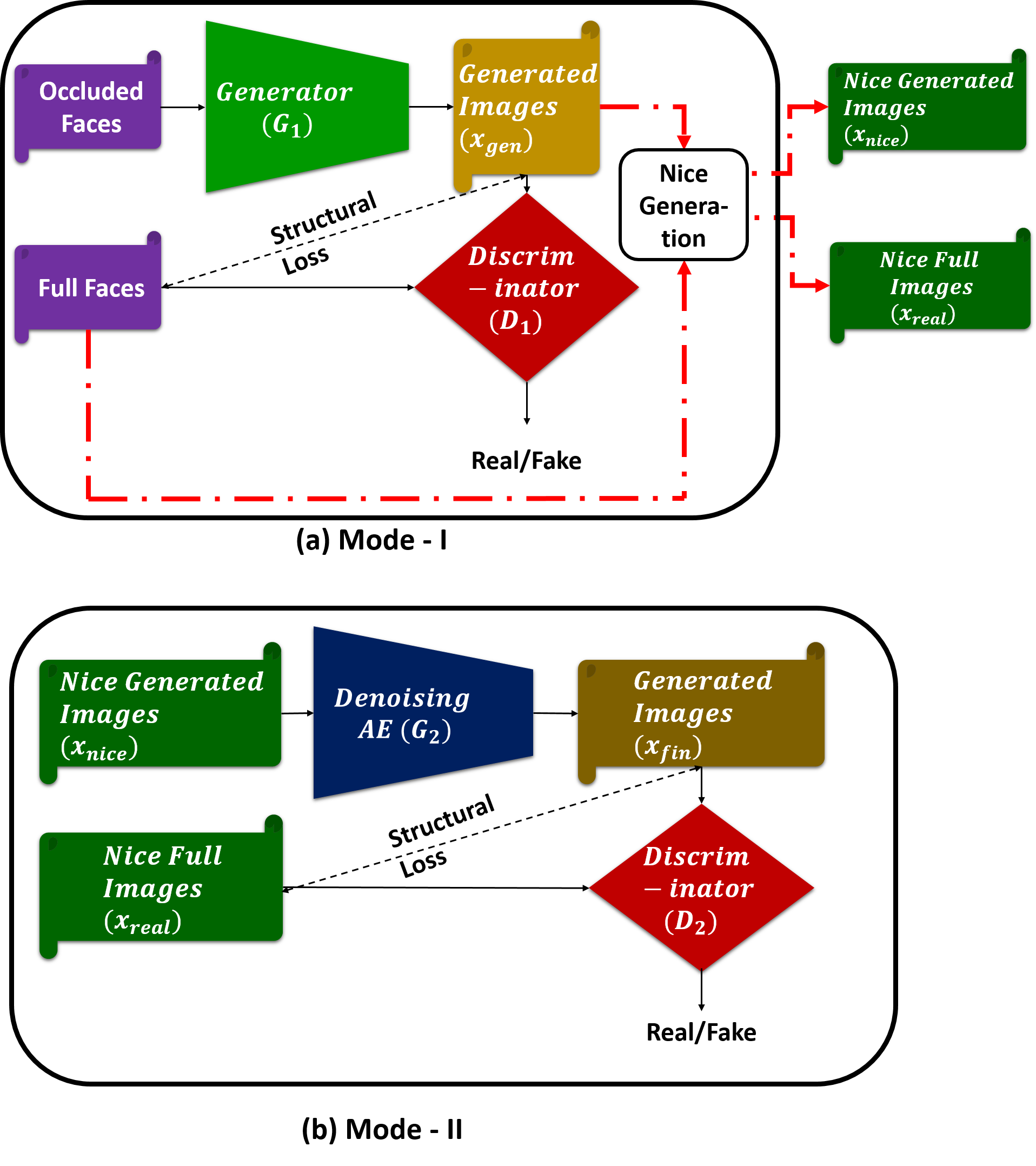}%
	\caption{The proposed SD-GAN architecture (best viewed in color), exhibiting two modes of operations (training): (a) Mode - I and (b) Mode - II.}
	\label{fig:sd_gan}
\end{figure*}

\begin{figure*}[!htbp]
	\centering
	\subfloat[Generator ($G_1$)]{{\includegraphics[width = 0.8\linewidth]{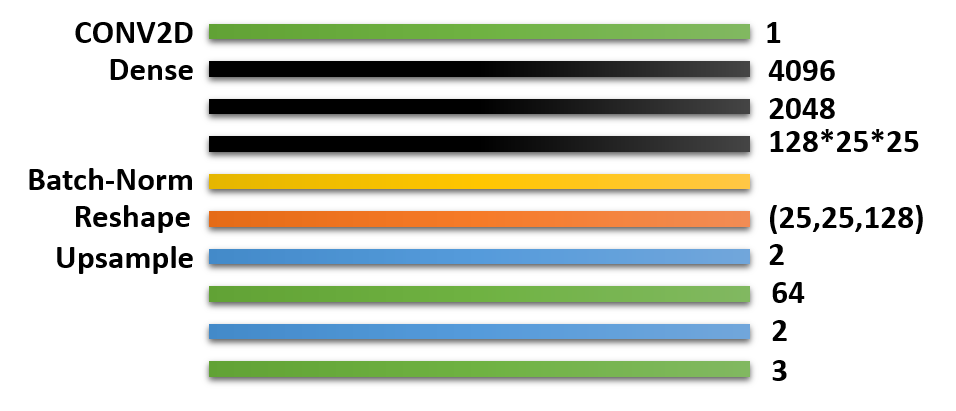}}}%
	\quad
	\subfloat[Discriminator ($D_1 / D_2$)]{{\includegraphics[width = 0.8\linewidth]{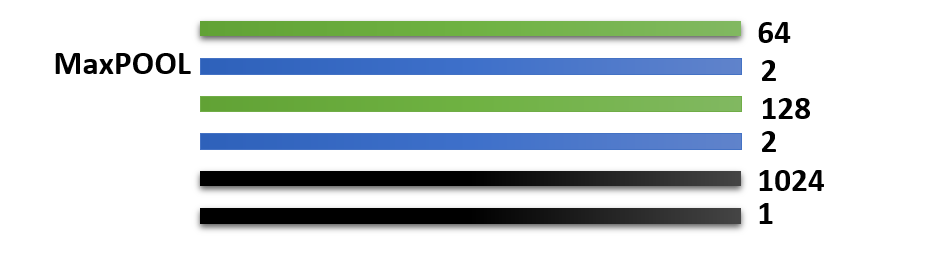}}}%
	\quad 
	\subfloat[Denoising AE ($G_2$)]{{\includegraphics[width = 0.8\linewidth]{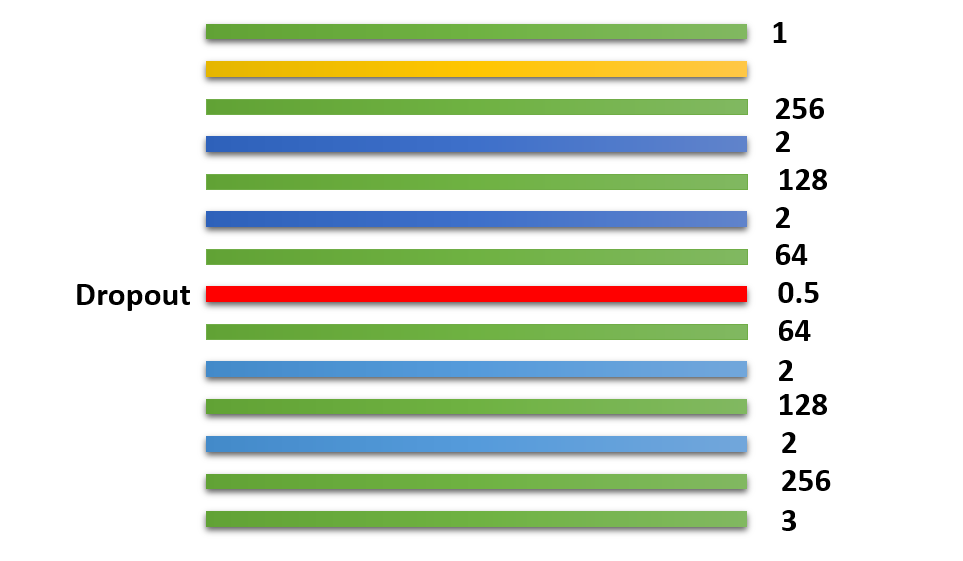}}}
	\caption{Architectural Details of the CNN-based Generator and Discriminator used in SD-GAN (best viewed in color).}
	\label{fig:net}
\end{figure*}

The proposed Structural and Denoising Generative Adversarial Network (SD-GAN) works in two-modes. Figures \ref{fig:sd_gan} \& \ref{fig:net} show the proposed architecture with structural details of SD-GAN, and descriptions for each of the modes of operation are described in the following sub-sections.

\subsection{Mode-I}
\label{sec:p1_sdg}
The Mode-I of SD-GAN is derived from DC-GAN \cite{radford2015unsupervised}, with a few variations in the input as well as in the training procedure (see section \ref{sec:train} for further details). The generator, $G_1$, is a deep-network (see figure \ref{fig:net}(a)) which takes the occluded faces as input, instead of the noise vector (as in DC-GAN) and generates (synthetic) faces to be fed to the discriminator $D_1$. $D_1$, similar to the discriminator network in DC-GAN (see figure \ref{fig:net}(b)), takes both the full real-world facial as well as $x_{gen}$ as inputs and attempts to discriminate between the real and generated (fake) images.

A "nice generation" module acts as an interface for selective data transfer between two modes of training. It takes fake images ($x_{gen}$) as input with a mini-batch of size $20$, and computes a loss function (see line 5 of algorithm \ref{algo:SDGAN}) to filter and create nice images ($x_{nice}$), when the loss is significantly low ($<0.01$). The corresponding full face images are also filtered as $x_{real}$ and given to mode-II of training. This is done under the assumption that $G_1$ has successfully fooled $D_1$ for the batch of images, when loss is low. Since $x_{nice}$ are often corrupted by noise, an operation of denoising is necessary as done by mode-II of operation.

\subsection{Mode-II}
\label{sec:p2_sdg}
The Mode-II is the denoising unit of our proposed architecture, compared to the Mode-I which preserves the structural identity of the face. To perform the task of denoising, a denoising auto-encoder (see section \ref{sec:DAE}) is used as the generator ($G_2$) in this mode of operation. For the CNN-based denoising auto-encoder (refer figure \ref{fig:net}(c)) proposed in this paper, the generated "nice" images ($x_{nice}$) obtained from Mode-I are taken as inputs. The discriminator ($D_2$), identical to $D_1$, takes as input $x_{real}$ images and performs adversarial training independently and exclusively. Though the input to Mode-II is given as output of Mode-I, the training and weight update of the model at Mode-II is independent of the training of Mode-I, \textit{i.e.} the gradients do not backpropagate into the model of Mode-I.


\section{Training SD-GAN}
\label{sec:train}

The bimodal SD-GAN model is trained using the proposed algorithm \ref{algo:SDGAN}. The procedure involves an end-to-end training of both the modes simultaneously. Each mode is trained using a procedure adopted from DC-GAN \cite{radford2015unsupervised}, with a structural loss induced for each mode, exclusively. The model is trained in \textit{Keras} with \textit{Tensorflow} backend \cite{abadi2016tensorflow}. A uniform mini-batch size of $20$ samples has been used throughout the training process, with gradient based optimization for weight update in the network. The following sub-sections detail the mode-wise training procedure, with the loss functions involved for weight update in the network (for all notations used hereafter, refer algorithm \ref{algo:SDGAN}).

\begin{algorithm}[!htbp]
	
	\SetAlgoLined
	\KwIn{Masked Face Image ($F_m$); Full Face ($F_f$)}
	\KwOut{Trained models $\rightarrow \{D_1, G_1, D_2, G_2\}$}
	$B$ := mini-batch from $F_m$ \& $F_f$\\
	$x_{nice}$ $\leftarrow$ []; 
	$x_{real}$ $\leftarrow$ [] \\
	\While{epoch $\le$ $100000$}{
		\ForEach{$F_m^B$ in $B$}{
			\tcp{$F_m^B$ $\in$ $F_m$ in batch $B$ \\
				$F_f^B$ $\in$ $F_f$ in batch $B$ }
			Compute $\mathcal{L}_{D_1}^{adv}(F_f^B, F_m^B)$ using equation \ref{eq:d1_loss} \& minimize\\
			$D_1$.trainable := \textbf{False}\\
			Compute $\mathcal{L}_{G_1}^{adv}(F_m^B, F_f^B)$ using equation \ref{eq:g1_loss} \& minimize\\
			$D_1$.trainable := \textbf{True}\\
			\If{$x_{loss} \le 0.01$}{
				$x_{nice}$ := append($x_{nice},x_{gen}$)\\
				$x_{real}$ := append($x_{real},F_f^B$)\\
			}
			\If{$x_{nice}$ is not empty}{
				$B_n$ := mini-batch from $x_{nice}$ \& $x_{real}$\\
				\ForEach{$N_m^B$ in $B_n$}{
					\tcp{$N_m^B$ $\in$ $x_{nice}$ in batch $B_n$ \\
						$O_f^B$ $\in$ $x_{real}$ in batch $B_n$ }
					Compute $\mathcal{L}_{D_2}^{adv}(O_f^B, N_m^B)$ using equation \ref{eq:d2_loss} \& minimize\\
					$D_2$.trainable := \textbf{False}\\
					Compute $\mathcal{L}_{G_2}^{adv}(N_m^B, O_f^B)$ using equation \ref{eq:g2_loss} \& minimize\\
					$D_2$.trainable := \textbf{True}\\}}
		}
		epoch := epoch+1
	}
	\tcc{D.trainable = FALSE indicates that the weights are frozen, and when TRUE weight update is performed using Backpropagation.}
	\caption{Overall training algorithm for SD-GAN}
	\label{algo:SDGAN}
\end{algorithm}

\subsection{Training for Mode-I}
\label{tr_p1}
The training process used for Mode-I is outlined in lines $4-12$ of algorithm \ref{algo:SDGAN}. The occluded images are given as inputs to $G_1$, to generate fake images matching the underlying true distribution of the full-facial images. The semi-supervised training procedure of SD-GAN involves a discriminator $D_1$ to distinguish between the real-world and generated images. The full-faces corresponding to each of the occluded faces in a batch, $B$, is fed to the discriminator as real images. The training of $D_1$ is based on the minimization of the binary cross-entropy loss ($\mathcal{L}_{bce}$) (see section \ref{sec:bce} for details), using the ADAM \cite{kingma2014adam} optimizer. Let, $x_{real}$ represent the set of full real-world face images and $x_{occ}$ be the occluded faces in a particular batch, while $D_1(x,y)$ represents the discriminator function with an input $x$ and a target label $y$ (set as $1$ for $x_{real}$ and $0$ for $x_{occ}$), and $G_1(x)$ depicts the generating function with the input $x$. The adversarial loss corresponding to $D_1$ can be written as:
\begin{equation}
\begin{split}
\mathcal{L}_{D_1}^{adv} (x_{real}, x_{occ}) = & \\ 
\mathcal{L}_{bce}(D_1(x_{real},y),\vec{1})+ &\mathcal{L}_{bce}(D_1(G_1(x_{occ}),y),\vec{0})\\
\end{split}	
\label{eq:d1_loss}
\end{equation}

Training the generator $G_1$ is essentially an optimization process executed using Stochastic Gradient Descent (SGD) \cite{amari1993backpropagation}, while freezing the weight update of $D_1$. The proposed structural loss (auxiliary) is induced at this stage of training. The adversarial loss for $G_1$ is:
\begin{equation}
\begin{split}
\mathcal{L}_{G_1}^{adv} (x_{occ}, x_{real}) & \\ =\mathcal{L}_{bce}(D_1(G_1(x_{occ}),y),\vec{1})+ & \mathcal{L}_{st}(x_{real}, G_1(x_{occ})) \\
\end{split}	
\label{eq:g1_loss}
\end{equation}
where, $\mathcal{L}_{st}$ is defined in equation \ref{eq:st}.

Minimization of these two criteria given by equations (\ref{eq:d1_loss}) and (\ref{eq:g1_loss}), makes $G_1$ outsmart (by cheating) $D_1$ upon reaching Nash equilibrium \cite{gibbons1992primer}, where $D_1$ believes that the images generated by $G_1$ is sampled from the true distribution.


\subsection{Training for Mode-II}
\label{sec:tr_p2}

The output images obtained from Mode-I are used in training for Mode-II in SD-GAN. Hence, these batch of "nice" images ($x_{nice}$) generated by $G_1$ are provided as inputs to Mode-II along with their corresponding (subject-wise) full-face images ($x_{real}$). Though, these images have their structural content partly preserved, they suffer from few degradation due to noise. To denoise these images, a denoising auto-encoder based generator model had been proposed in this paper. Lines $14-20$ in algorithm \ref{algo:SDGAN} outlines mode-II of training. The Discriminator $D_2$ comprises of a similar adversarial loss as in $D_1$, given as:
\begin{equation}
\begin{split}
\mathcal{L}_{D_2}^{adv} (x_{real}, x_{nice}) = &\\ 
\mathcal{L}_{bce}(D_2(x_{real},y),\vec{1})+ &\mathcal{L}_{bce}(D_2(G_2(x_{occ}),y),\vec{0})\\
\end{split}
\label{eq:d2_loss}
\end{equation}

The denoising auto-encoder training of $G_2$ is incremental, in a sense that the number of training samples increases as the $G_1$ becomes stronger. The instability issues \cite{goodfellow2014generative} prevalent in training is taken care by over-training the weaker of the two to reach the equilibrium point. The adversarial loss incurred at this phase mainly deals with closing the gap between the distributions of the real and the generated (fake) samples. The adversarial loss at $G_2$ is given by:
\begin{equation}
\begin{split}
\mathcal{L}_{G_2}^{adv} (x_{nice}, x_{real}) = &\\ 
\mathcal{L}_{bce}(D_2(G_2(x_{occ}),y),\vec{1})+ &\mathcal{L}_{aux}(x_{real}, G_1(x_{occ})) \\
\end{split}
\label{eq:g2_loss}
\end{equation}
where,\\ 
$\mathcal{L}_{aux} = \triangle \big(\mathcal{L}_{st}(x_{real}, G_1(x_{occ})), \mathcal{L}_{st}(x_{real}, G_2(x_{nice}))\big)$, and $\triangle$ being the difference operator.

Minimization of $\mathcal{L}_{G_2}^{adv}$ reduces the gap in structural and pixel-values between the generated (fake) and true samples, which also reduces the noise in the generated samples.

The use of Mode-II of training along with Mode-I (done independently) reduces the overall time for training ($\sim10^2$ folds, considering the number of epochs) compared to a recent state-of-the-art technique \cite{li2017generative} used for the task at hand.  
\section{Results and Performance Analysis}
\label{sec:res}
This section first describes the datasets used, then gives the quantitative measures used to show the effectiveness of our proposed model for face completion and FR, compared with a few state-of-the-art techniques.
\subsection{Datasets}
\label{sec:ds}
Experimentations are carried on three datasets: (a) AR dataset \cite{martinez1998ar}, (b) Celeb-A dataset \cite{liu2015faceattributes}, and (c) multi-PIE \cite{gross2010multi}; each is briefly described below.
\subsubsection{AR Database}
\label{sec:ar}
\begin{figure}[!htbp]
	\centering
	\subfloat[]{{\includegraphics[width = 0.18\linewidth, height = 25mm]{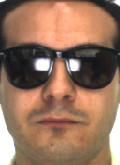}}}%
	\quad
	\subfloat[]{{\includegraphics[width = 0.18\linewidth, height = 25mm]{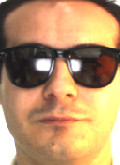}}}%
	\quad 
	\subfloat[]{{\includegraphics[width = 0.18\linewidth, height = 25mm]{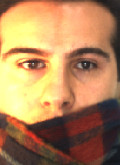}}}
	\quad
	\subfloat[]{{\includegraphics[width = 0.18\linewidth, height = 25mm]{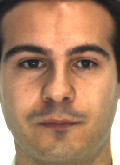}}}%
	\quad\\
	\subfloat[]{{\includegraphics[width = 0.18\linewidth, height = 25mm]{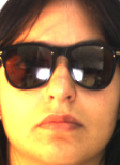}}}
	\quad
	\subfloat[]{{\includegraphics[width = 0.18\linewidth, height = 25mm]{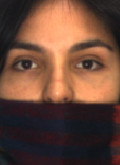}}}
	\quad
	\subfloat[]{{\includegraphics[width = 0.18\linewidth, height = 25mm]{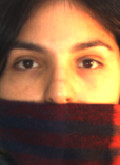}}}
	\quad
	\subfloat[]{{\includegraphics[width = 0.18\linewidth, height = 25mm]{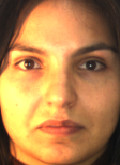}}}
	\caption{Two face image examples from AR database (one in each row) with different levels of occlusions and illumination variations. Images in \{(a), (b), (e)\} $\in$ \textbf{AR1}; while \{(c), (f), (g)\} $\in$ \textbf{AR2}; and \{(d), (h)\} are the full face images (best viewed in color).}
	\label{fig:ar}
\end{figure}

The AR database \cite{martinez1998ar} consists of face images which contain real-world occlusions. The database consists of $136$ subjects with varying illumination conditions and expressions. For our study, we consider those images which are near-frontal and have minimal expression variations (see figure \ref{fig:ar} for samples). Two variations of occlusions are available in the database, \textit{viz.} the sunglasses and scarf on the face, which prevents the faces to be reconstructed using symmetric transformations from the other half of the face. For our experimentations, the dataset has been divided into 2 subsets: \textbf{AR1}, the images with sunglasses and \textbf{AR2}, those with scarfs. A data partition as $60:20:20$ ratio is maintained uniformly for training, validation, and testing throughout the set of the experimentations. The subjects used for training and validation are never used for testing.

\begin{figure}[!htbp]
	\centering
	{\includegraphics[width = \linewidth]{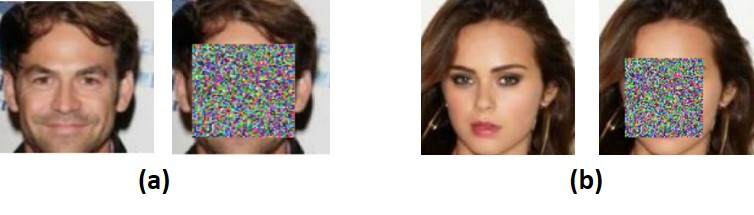}}
	\caption{Examples from CelebA database, showing two subjects with synthetic occlusions (best viewed in color).}
	\label{fig:ca}
\end{figure}

\subsubsection{Celeb-A Database}
\label{sec:celeba}
The CelebA \cite{liu2015faceattributes} dataset consists of 202,599 face images. Each face image is cropped, roughly aligned by the position of two eyes, and rescaled to $100\times100\times3$ pixels. The standard benchmark split with 162,770 images for training, 19,867 for validation and 19,962 for testing, has been followed for experimentation. A mask of size $50 \times 50$ pixels covers the face (see figure \ref{fig:ca} for samples) at random locations, as described in \cite{li2017generative}.

\subsubsection{Multi-PIE dataset}
\label{sec:mpa}
The CMU Multi-PIE database \cite{gross2010multi} consists of 755,370 images shot in 4 different
sessions from 337 subjects. The images in the dataset are split up into training, validation and
test set. The training set is composed of all individuals in non-frontal pose (except those used for validation and testing) at the generator, while the size of the validation (64 identities at a pose of $90^\circ$) and test sets (65 identities at a pose of $90^\circ$) are almost identical. We consider the images taken in session 1, with the probe images taken at $90^\circ$ pose.
\subsection{Evaluation metrics}
\label{sec:metric}
Along with the visual results shown in section \ref{sec:pa_gen} we perform quantitative evaluation of the proposed model for the two datasets under test. Firstly, we use the peak-signal-to-noise-ratio (PSNR) value, which captures the difference in the pixel values of the two images. PSNR (higher the better) is defined as:

\begin{equation}
\begin{split}
MSE(x_{fin},x_{real}) &= \frac{1}{mn}\sum_{i=1}^{m-1}\sum_{i=1}^{n-1}\big[x_{gen}(i,j)-x_{real}(i,j)\big]^2 \\
PSNR &= 10 \cdot \log_{10} \bigg(\frac{MAX_{x_{fin}}^2}{MSE}\bigg)
\end{split}	
\end{equation}
where, $x_{fin}$ is the output (generated) image and $x_{real}$ is the reference (ground-truth, GT) image.

Secondly, SSIM index (refer equation \ref{eq:ssim}) is used for quantifying the generated results, which estimates the holistic similarity between two images. Finally, we also use the identity distances measured by the OpenFace toolbox \cite{amos2016openface} to determine the high-level semantic similarity of two faces.

\subsection{Performance Analysis for generation of full facial images}
\label{sec:pa_gen}

\begin{figure}[!htbp]
	\centering
	\includegraphics[width=\textwidth]{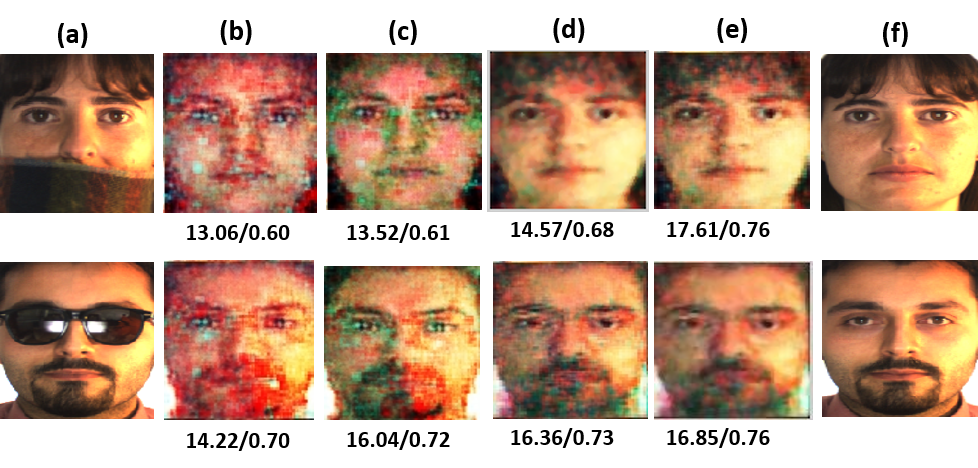}
	\caption{Results for image generation from two different sets of occlusions, \textit{viz.}, \textbf{AR2} and \textbf{AR1} (arranged row-wise) present in AR, by SD-GAN: (a) the input occluded image, (b) output of $G_1$ using $\mathcal{L}_{bce}$, (c) output of $G_1$ using $\mathcal{L}_{bce} + \mathcal{L}_{ssim}$, (d) output of $G_1$ at Phase-I, (e) output of $G_2$ at Phase-II, (f) Ground-truth (GT). The values below each image from (b)-(e) give the (PSNR/SSIM) values of the images compared to the expected output (GT).}
	\label{fig:res_ar}
\end{figure}

A few examples of generation of the full facial images from occluded faces are shown in figure \ref{fig:res_ar} under two different scenarios of the proposed method. The column (b) depicts the output of DC-GAN \cite{radford2015unsupervised}, while the results progressively becomes better as we move towards the right, showing the effectiveness of the auxiliary losses proposed in this paper. The significant improvement in the image quality measure shown by our model in (e) as compared to (d) (see table \ref{tab:quant} for quantitative measures showing similar trends) strengthens our claim for the introduction of Mode-II for denoising the output of Mode-I.

\begin{figure}[!htbp]
	\centering
	\includegraphics[width=0.7\textwidth]{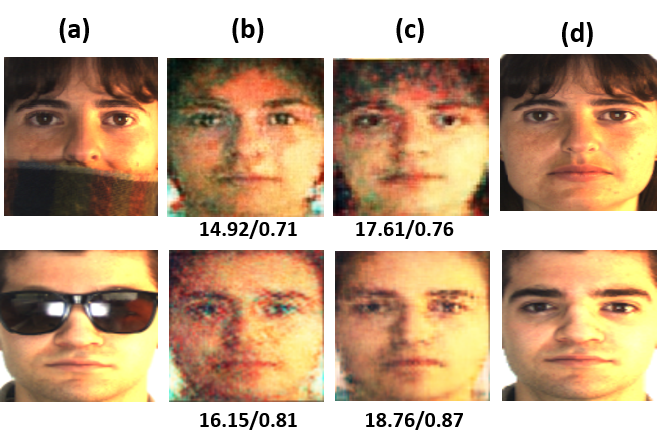}
	\caption{Results for image generation from two different methods: (a) occluded images (one each from \textbf{AR2} (\textit{Top-row}) and \textbf{AR1} (\textit{Bottom-Row})), (b) Images generated by GFC \cite{li2017generative} without post-processing, (c) Images generated by SD-GAN, (d) expected output. The values below each image gives the (PSNR/SSIM) values of the images compared to the expected output.}
	\label{fig:comp}
\end{figure}

\begin{figure}[!htbp]
	\centering
	\includegraphics[width=0.75\textwidth]{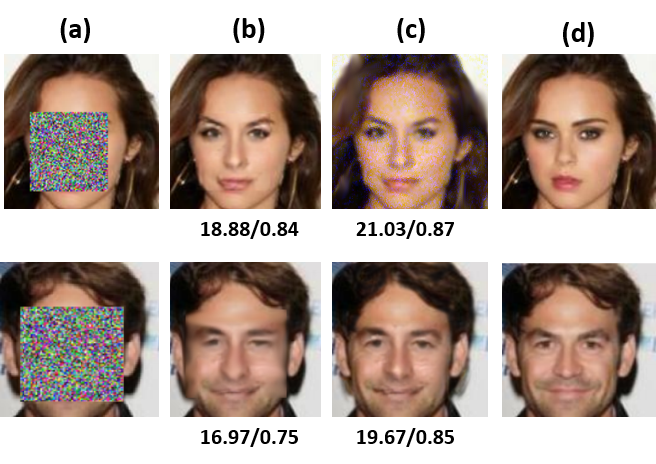}
	\caption{Results for image generation from two different methods: (a) occluded images (from Celeb-A dataset \cite{liu2015faceattributes}), (b) Images generated by GFC \cite{li2017generative} without post-processing, (c) Images generated by SD-GAN, (d) expected output. The values below each image gives the (PSNR/SSIM) values of the images compared to the expected output.}
	\label{fig:comp1}
\end{figure}

Both the quantitative as well as the qualitative measures are compared with a recent state-of-the-art technique. GFC \cite{li2017generative} uses face parsing as well as Poisson Blending \cite{perez2003poisson} as post-processing techniques to generate facial parts under occlusion. Graph Laplacian (GL) based methods \cite{deng2011graph} also attempts to solve the problem. The quantitative results evaluating the quality of the images are given in table \ref{tab:quant}. Our proposed SD-GAN (referred as 'SDG' in tables) outperforms all other techniques based on PSNR values, whereas in case of the holistic measure (SSIM), the nearest competing method GFC, also a GAN based deep model with post-processing techniques, matches our performance in a few cases and even marginally outperforms our proposed technique in only one case. Qualitative experiments also reveal that without the post-processing technique, GFC fails to match the performance of our proposed technique in both the datasets, for which our method is a clear winner, as shown in figures \ref{fig:comp} \& \ref{fig:comp1}. The values at the bottom of the images in columns (b) \& (c) in figures \ref{fig:comp} \& \ref{fig:comp1}, reveal the superiority of our proposed SD-GAN, based on the PSNR/SSIM values on the four exemplar images.
\begin{table}[!htbp]
	\centering
	\caption{Quantitative values (averaged over the whole dataset) for different face images generated following the protocol, as in columns (b)-(d) of figure \ref{fig:res_ar} for the AR dataset, compared with state-of-the-art techniques.}
	\begin{tabular}{p{1.5cm} p{1cm} p{1cm} p{1cm} p{1cm} p{1cm} p{1cm}}
		\hline\hline
		\multicolumn{7}{c}{\textbf{\textit{Average PSNR Values}} (higher the better)}\\\hline
		&\textbf{(b)} & \textbf{(c)} & \textbf{(d)} & \textbf{GL} & \textbf{GFC} & \textbf{SDG} \\\hline
		\textit{AR1} & 12.15 & 13.27 & 15.62 & 13.48 & 15.83 & \textbf{18.43} \\
		\textit{AR2} & 11.92 & 12.58 & 14.87 & 11.78 & 13.84 & \textbf{17.68} \\
		\textit{CelebA} & 12.31 & 12.86 & 16.82 & 9.43 & 18.30 & \textbf{18.61} \\\hline\hline
		
		\multicolumn{7}{c}{\textbf{\textit{Average SSIM Indices}} (higher the better)}\\\hline
		&\textbf{(b)} & \textbf{(c)} & \textbf{(d)} & \textbf{GL} & \textbf{GFC} & \textbf{SDG} \\\hline
		\textit{AR1} & 0.67 & 0.70 & 0.70 & 0.65 & \textbf{0.77} & \textbf{0.77} \\
		\textit{AR2} & 0.59 & 0.65 & 0.70 & 0.54 & 0.73 & \textbf{0.76} \\
		\textit{CelebA} & 0.68 & 0.71 & 0.73 & 0.67 & \textbf{0.76} & \textbf{0.76} \\\hline\hline
		
		\multicolumn{7}{c}{\textbf{\textit{Average Identity distances}} (lower the better)}\\\hline
		&\textbf{(b)} & \textbf{(c)} & \textbf{(d)} & \textbf{GL} & \textbf{GFC} & \textbf{SDG} \\\hline
		\textit{AR1} & 0.64 & 0.61 & 0.52 & 0.52 & 0.48 & \textbf{0.47} \\
		\textit{AR2} & 0.75 & 0.72 & 0.59 & 0.67 & \textbf{0.56} & \textbf{0.56} \\
		\textit{CelebA} & 0.68 & 0.62 & 0.59 & 0.61 & \textbf{0.55} & 0.57 \\\hline\hline
	\end{tabular}
	\label{tab:quant}
\end{table}

\subsection{Performance boost in Face Recognition}
\label{sec:perf_ver}
Face Recognition (FR) systems underperform when the faces are occluded. Our proposed SD-GAN reconstructs a full-face when presented with a occluded face, which facilitates efficient performance for FR. Performances of several recent shallow learning techniques, \textit{viz.} LSM \cite{hwang2003reconstruction}, RPCA \cite{wright2009robustpca}, GL \cite{deng2011graph} have been compared with our proposed and GFC \cite{li2017generative} methods for generation of the faces, evaluated using state-of-the-art benchmark FR systems, like PCA \cite{turk1991face}, Gabor \cite{lei2008gabor}, LPP \cite{he2004locality}, Sparse Representation (SR) \cite{wright2009robust} and VGG \cite{parkhi2015deep}. The results in table \ref{tab:ar_rec} show the rank-1 accuracies for AR1 and AR2 datasets, where our proposed model (SD-GAN) outperforms all other methods, indicating that it must be capable of generating discriminative parts of the face better than the other competing methods. Interpret the values in the table \ref{tab:ar_rec} as performances for FR, for images generated by the methods mentioned at the top of each column, while the FR methods appear at the left of each row. Observe the huge jump in performance from the statistical methods to the GAN based methods, indicating the power of the GAN based techniques for overcoming occluded faces, specifically when applied for FR applications. 
\begin{table}[!htbp]
	\centering
	\caption{Rank-1 Recognition rates (in \%) exhibiting a higher performance for Face Recognition by SD-GAN (SDG), compared with several state-of-the-art shallow and deep learning techniques on AR Dataset. The results in bold demarcates the best performance (row-wise).}
	\begin{tabular}{p{1cm} p{1cm} p{1cm} p{1cm} p{1cm} p{1cm} p{1cm}}
		\hline\hline
		\multicolumn{7}{c}{\textbf{AR1}: Recognition of faces with sunglasses}\\
		\hline
		&\textbf{Occ.}&\textbf{LSM}&\textbf{RPCA}&\textbf{GL}&\textbf{GFC}&\textbf{SDG}\\\hline
		\textit{PCA }& 52.6 & 61.4 & 64.2 & 70.0 & 82.9 & \textbf{89.7} \\
		\textit{GPCA} & 67.5 & 73.3 & 71.6 & 76.6 & 88.4 & \textbf{93.3} \\
		\textit{LPP} & 53.4 & 45.7 & 61.4 & 59.0 & 83.5 & \textbf{90.1} \\
		\textit{SR} & 58.4 & 59.2 & 57.3 & 60.6 & 85.7 & \textbf{91.6} \\
		\textit{VGG} & 84.2 & 85.4 & 84.5 & 87.9 & 91.7 & \textbf{96.8} \\\hline\hline
		\multicolumn{7}{c}{\textbf{AR2}: Recognition of faces with scarf}\\
		\hline
		&\textbf{Occ.}&\textbf{LSM}&\textbf{RPCA}&\textbf{GL}&\textbf{GFC}&\textbf{SDG}\\\hline
		\textit{PCA }& 15.7 & 37.5 & 32.2 & 40.8 & 72.6 & \textbf{79.4} \\
		\textit{GPCA} & 55.1 & 56.2 & 54.0 & 60.9 & 80.3 & \textbf{88.6} \\
		\textit{LPP} & 34.4 & 43.0 & 38.3 & 47.1 & 75.9 & \textbf{81.2} \\
		\textit{SR} & 45.2 & 51.8 & 47.7 & 56.7 & 79.8 & \textbf{86.8} \\
		\textit{VGG} & 72.3 & 75.9 & 79.6 & 83.5 & 89.9 & \textbf{92.6} \\\hline\hline
	\end{tabular}
	\label{tab:ar_rec}
\end{table}

An extension of Linear Discriminant Analysis (LDA) \cite{gunther2016face} to the two color channels I-chrominance and the Red channel (LDA-IR) is described in \cite{gunther2016face}. Inter-Session Variability (ISV) \cite{gunther2016face} modeling is a technique that has been successfully employed for face verification, which does not have occluded images during training. The rank-1 recognition rates of the VGG+SD-GAN (VGG is used as a classifier with SDG as the generator), when compared with these two state-of-the-art techniques, LDA-IR and ISV, are much higher for the AR database, as reported in table \ref{tab:gun}.

\begin{table}[!htbp]
	\centering
	\caption{Rank-1 Recognition rates for end-to-end system for occluded face recognition. Higher values are better.}
	\begin{tabular}{p{2cm} p{2cm} p{2.5cm} p{2cm} }
		\hline\hline
		\textbf{Dataset}& \textbf{ISV \cite{gunther2016face}} & \textbf{LDA-IR \cite{gunther2016face}} & \textbf{VGG+SDG}\\\hline 
		AR1 & 45.13 & 62.59 & \textbf{96.82} \\\hline
		AR2 & 39.81 & 57.44 & \textbf{92.64} \\\hline\hline
	\end{tabular}
	\label{tab:gun}
\end{table}

\subsection{Analysis of training time of SD-GAN, compared to GFC \cite{li2017generative}}
\label{sec:time}
\begin{figure}[!htbp]
	\centering
	\includegraphics[width=0.7\linewidth]{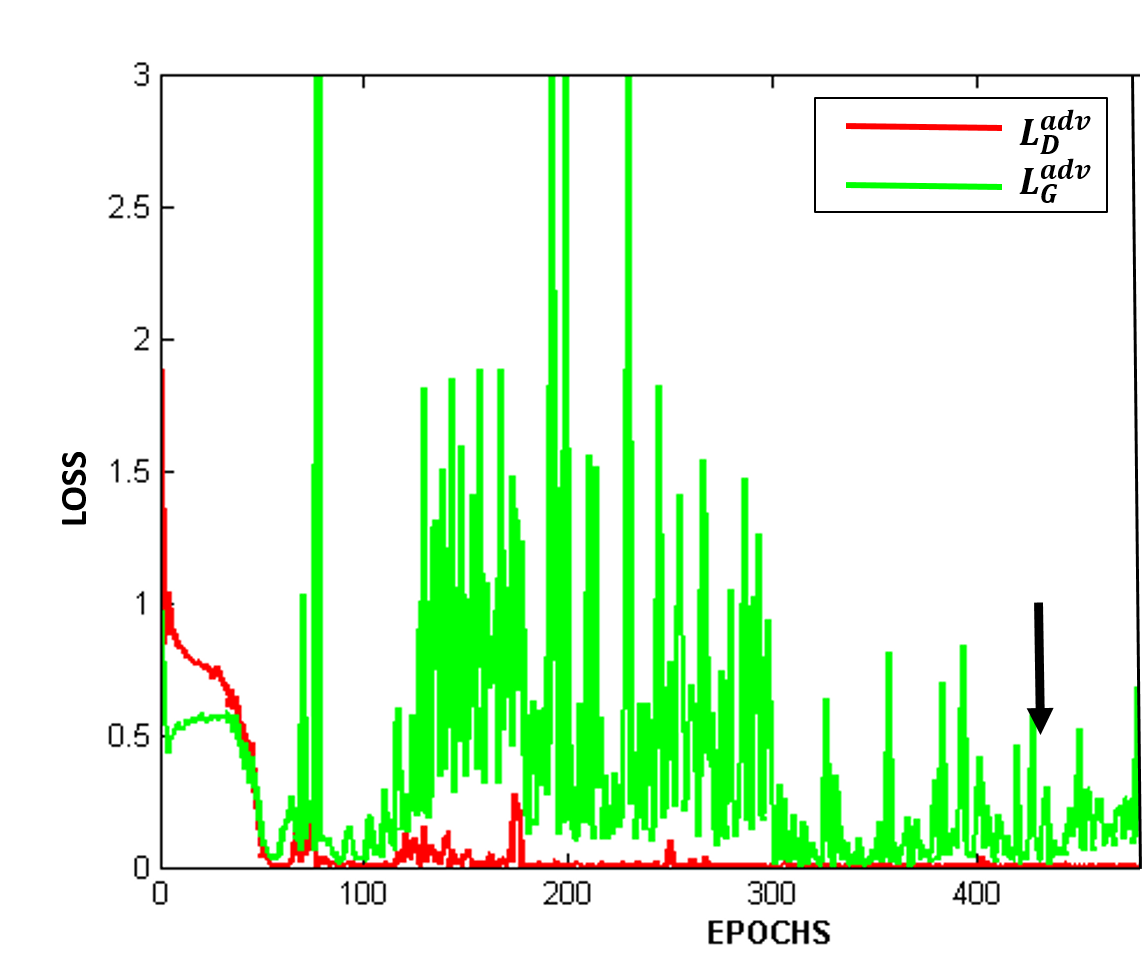}
	\caption{Graphs showing the discriminator and generator loss functions during training.}
	\label{fig:eq_graph}
\end{figure}

All experiments are performed on a dual GPU machine with dual Nvidia TITAN X, with 64 GB RAM and Intel core i7 4790K processor. The training for both the models are performed using $Keras$ with $Tensorflow$ backend. The training times are tabulated in table \ref{tab:time}, which shows that the SD-GAN is faster than GFC, since it converges near a Nash equilibrium (see arrow on graph in figure \ref{fig:eq_graph} for details) in lesser number of epochs as compared to GFC.
\begin{table}[!htbp]
	\centering
	\caption{Comparison of training times of SD-GAN and GFC. Lower value is better.}
	\begin{tabular}{p{1cm} p{2cm} p{2cm} p{0.1cm} p{2cm} p{2cm}}
		\hline\hline
		&\multicolumn{2}{c}{AR Face Database} & &\multicolumn{2}{c}{Celeb-A Database}\\
		\cline{2-3} \cline{5-6}
		&\textbf{\#epochs}&\textbf{mins/epoch}& &\textbf{\#epochs}&\textbf{mins/epoch}\\\hline
		GFC & 30K & 8 & & 20K & 25 \\
		SDG & \textbf{550} & \textbf{3} && \textbf{500} & \textbf{12} \\\hline\hline
	\end{tabular}
	\label{tab:time}
\end{table}

\subsection{Results on the Multi-PIE dataset}
\begin{figure}[!htbp]
	\centering
	\includegraphics[width=0.8\textwidth]{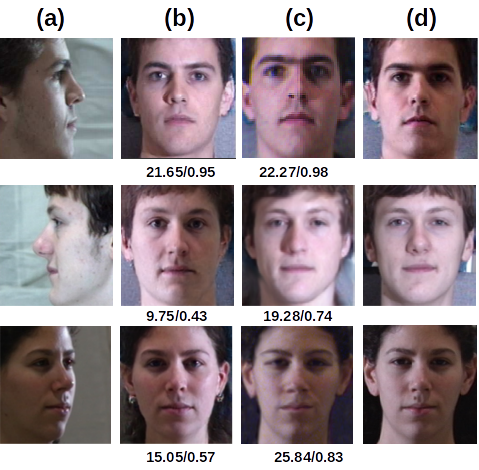}
	\caption{Results for image generation from two different methods: (a) Images at different poses (obtained from Multi-Pie dataset \cite{gross2010multi}, with left- (\textit{top-row}) \& right-looking (\textit{Middle-row}) profiles at $90^\circ$; and a face image at $60^\circ$ pose (\textit{Bottom-row})) used for testing, (b) Image generated by TPG \cite{huang2017beyond}, (c) Image generated by SDG, (d) expected output (ground-truth).The values below each image gives the (PSNR/SSIM) values of the image compared to the expected (target) output.}
	\label{fig:pose}
\end{figure}

In order to evaluate our proposed algorithm on pose-variations of the face images producing self-occlusions, we performed experimentations on MultiPIE dataset, which has 750000+ images, at different poses. Self-occlusion of faces occur due to off-frontal and out-of-plane rotation variations in pose. For evaluating performance using rank-1 recognition rates, we follow the protocol from [33], and only images from session one are used. Results are given in table \ref{tab:pose}. All images used for testing and validation have $90^\circ$ pose. Few results shown in figure \ref{fig:pose} display the superiority of our method over TP-GAN (TPG) \cite{huang2017beyond}, both qualitatively as well as with quantitative measures in terms the SSIM/PSNR values. Observe the sample at the last row of figure \ref{fig:pose}, which shows a non-frontal (not side profile view) query face. In this case, our result in (c) has produced an exact illumination variation as that in GT (d), whereas the process of  \cite{huang2017beyond} in (b) produces exactly the opposite (mirror-like image) while producing a sharper contrast (unnecessarily, in general) than that in GT. Also, observe intriguingly the presence of ear-rings (appears non-identical ones) in the output of \cite{huang2017beyond}, not present in GT and our output in (c). The proposed system intrinsically exploits the symmetric nature of the face, helping to generate images with appropriate illumination variations at high-resolution with desired quality as in GT.


\begin{table}[!htbp]
	\centering
	\caption{Comparison of Rank-1 recognition rate for Multi-PIE dataset, with faces at $90^\circ$ pose (best values are in bold).}
	\begin{tabular}{p{5cm} p{1.7cm} p{1.2cm} }
		\hline\hline
		\textbf{Criteria} & \textbf{TPG \cite{huang2017beyond}} & \textbf{SDG}\\\hline 
		Rank-1 Recognition Rate (\%) & 64.03 & \textbf{65.19}\\\hline
		PSNR & 12.26 & \textbf{19.84} \\\hline
		SSIM & 0.59 & \textbf{0.66} \\\hline\hline
	\end{tabular}
	\label{tab:pose}
\end{table}

\section{Conclusion}
\label{sec:conc}
The proposed SD-GAN model uses end-to-end training for reconstruction of occluded parts of the face. The proposed technique does not rely on any post-processing technique for semantic correction of the faces. Thus, this module may be used as pre-processing for any FR system, in cases where faces are occluded. A faster training time is ensured in this model, based on the Nash Equilibrium. The qualitative and the quantitative results discussed above confirm the superiority of our proposed model. Misalignment of faces may lead to distortions as happens in all reconstruction techniques. In order to generate better quality photo-realistic images for AR and LFW datasets, the dual pathway technique proposed in \cite{huang2017beyond} can be used as a post-processing stage following our SD-GAN.  

\section*{References}
{\small
	\bibliographystyle{elsarticle-num}
	\bibliography{egbib}

\begin{thebibliography}{10}
\expandafter\ifx\csname url\endcsname\relax
  \def\url#1{\texttt{#1}}\fi
\expandafter\ifx\csname urlprefix\endcsname\relax\def\urlprefix{URL }\fi
\expandafter\ifx\csname href\endcsname\relax
  \def\href#1#2{#2} \def\path#1{#1}\fi

\bibitem{goodfellow2014generative}
I.~Goodfellow, J.~Pouget-Abadie, M.~Mirza, B.~Xu, D.~Warde-Farley, S.~Ozair,
  A.~Courville, Y.~Bengio, Generative adversarial nets, in: Advances in Neural
  Information Processing Systems (NIPS), 2014, pp. 2672--2680.

\bibitem{barnes2009patchmatch}
C.~Barnes, E.~Shechtman, A.~Finkelstein, D.~B. Goldman, Patchmatch: A
  randomized correspondence algorithm for structural image editing, ACM Trans.
  Graph. 28~(3) (2009) 24--1.

\bibitem{huang2014image}
J.-B. Huang, S.~B. Kang, N.~Ahuja, J.~Kopf, Image completion using planar
  structure guidance, ACM Transactions on Graphics (TOG) 33~(4) (2014) 129.

\bibitem{wright2009robust}
J.~Wright, A.~Y. Yang, A.~Ganesh, S.~S. Sastry, Y.~Ma, Robust face recognition
  via sparse representation, IEEE Transactions on Pattern Analysis and Machine
  Intelligence 31~(2) (2009) 210--227.

\bibitem{ren2015shepard}
J.~S. Ren, L.~Xu, Q.~Yan, W.~Sun, Shepard convolutional neural networks, in:
  Advances in Neural Information Processing Systems, 2015, pp. 901--909.

\bibitem{li2017generative}
Y.~Li, S.~Liu, J.~Yang, M.-H. Yang, Generative face completion, IEEE Conference
  on Computer Vision and Pattern Recognition (CVPR).

\bibitem{martinez1998ar}
A.~Martinez, R.~Benavente, The {AR} face database, CVC Tech. Report (1998) 24.

\bibitem{wang2004image}
Z.~Wang, A.~C. Bovik, H.~R. Sheikh, E.~P. Simoncelli, Image quality assessment:
  from error visibility to structural similarity, IEEE Transactions on Image
  Processing (TIP) 13~(4) (2004) 600--612.

\bibitem{krizhevsky2012imagenet}
A.~Krizhevsky, I.~Sutskever, G.~E. Hinton, Imagenet classification with deep
  convolutional neural networks, in: Advances in Neural Information Processing
  Systems (NIPS), 2012, pp. 1097--1105.

\bibitem{vincent2008extracting}
P.~Vincent, H.~Larochelle, Y.~Bengio, P.-A. Manzagol, Extracting and composing
  robust features with denoising autoencoders, in: Int'l Conference on Machine
  learning (ICML), 2008.

\bibitem{urgen1992learning}
J.~Schmidhuber, Learning factorial codes by predictability minimization, Neural
  Computation 4~(6) (1992) 863--879.

\bibitem{nash1950equilibrium}
J.~F. Nash, et~al., Equilibrium points in n-person games, Proceedings of the
  National Academy of Sciences 36~(1) (1950) 48--49.

\bibitem{bengio2007greedy}
Y.~Bengio, P.~Lamblin, D.~Popovici, H.~Larochelle, Greedy layer-wise training
  of deep networks, in: Advances in Neural Information Processing Systems
  (NIPS), 2007, pp. 153--160.

\bibitem{lee2012mmse}
C.~Lee, C.~Lee, C.-S. Kim, An {MMSE} approach to nonlocal image denoising:
  Theory and practical implementation, Journal of Visual Communication and
  Image Representation 23~(3) (2012) 476--490.

\bibitem{johnson2006stephen}
S.~Johnson, Stephen Johnson on digital photography, O'Reilly Media, Inc., 2006.

\bibitem{radford2015unsupervised}
A.~Radford, L.~Metz, S.~Chintala, Unsupervised representation learning with
  deep convolutional generative adversarial networks, in: International
  Conference on Learning Representations, 2015.

\bibitem{abadi2016tensorflow}
M.~Abadi, P.~Barham, J.~Chen, Z.~Chen, A.~Davis, J.~Dean, M.~Devin,
  S.~Ghemawat, G.~Irving, M.~Isard, et~al., Tensorflow: A system for
  large-scale machine learning., in: OSDI, Vol.~16, 2016, pp. 265--283.

\bibitem{kingma2014adam}
D.~P. Kingma, J.~Ba, Adam: A method for stochastic optimization, in:
  Proceedings of the 3rd International Conference on Learning Representations
  (ICLR), 2014.

\bibitem{amari1993backpropagation}
S.-i. Amari, Backpropagation and stochastic gradient descent method,
  Neurocomputing 5~(4) (1993) 185--196.

\bibitem{gibbons1992primer}
R.~Gibbons, A primer in game theory, Harvester Wheatsheaf, 1992.

\bibitem{liu2015faceattributes}
Z.~Liu, P.~Luo, X.~Wang, X.~Tang, Deep learning face attributes in the wild,
  in: Proceedings of International Conference on Computer Vision (ICCV), 2015.

\bibitem{gross2010multi}
R.~Gross, I.~Matthews, J.~Cohn, T.~Kanade, S.~Baker, Multi-pie, Image and
  Vision Computing (IVC) 28~(5) (2010) 807--813.

\bibitem{amos2016openface}
B.~Amos, B.~Ludwiczuk, M.~Satyanarayanan, Openface: A general-purpose face
  recognition library with mobile applications, Tech. rep., CMU-CS-16-118, CMU
  School of Computer Science (2016).

\bibitem{perez2003poisson}
P.~P{\'e}rez, M.~Gangnet, A.~Blake, Poisson image editing, in: ACM Transactions
  on graphics (TOG), Vol.~22, ACM, 2003, pp. 313--318.

\bibitem{deng2011graph}
Y.~Deng, Q.~Dai, Z.~Zhang, Graph laplace for occluded face completion and
  recognition, IEEE Transactions on Image Processing (TIP) 20~(8) (2011)
  2329--2338.

\bibitem{hwang2003reconstruction}
B.-W. Hwang, S.-W. Lee, Reconstruction of partially damaged face images based
  on a morphable face model, IEEE Transactions on Pattern Analysis and Machine
  Intelligence 25~(3) (2003) 365--372.

\bibitem{wright2009robustpca}
J.~Wright, A.~Ganesh, S.~Rao, Y.~Peng, Y.~Ma, Robust principal component
  analysis: Exact recovery of corrupted low-rank matrices via convex
  optimization, in: Advances in Neural Information Processing Systems, 2009,
  pp. 2080--2088.

\bibitem{turk1991face}
M.~A. Turk, A.~P. Pentland, Face recognition using eigenfaces, in: IEEE
  Computer Society Conference on Computer Vision and Pattern Recognition
  (CVPR), IEEE, 1991, pp. 586--591.

\bibitem{lei2008gabor}
Z.~Lei, S.~Liao, R.~He, M.~Pietikainen, S.~Z. Li, Gabor volume based local
  binary pattern for face representation and recognition, in: 8th IEEE
  International Conference on Automatic Face \& Gesture Recognition (FG), IEEE,
  2008, pp. 1--6.

\bibitem{he2004locality}
X.~He, P.~Niyogi, Locality preserving projections, in: Advances in Neural
  Information Processing Systems (NIPS), 2004, pp. 153--160.

\bibitem{parkhi2015deep}
O.~M. Parkhi, A.~Vedaldi, A.~Zisserman, Deep face recognition, in: British
  Machine Vision Conference (BMVC), Vol.~1, 2015, p.~6.

\bibitem{gunther2016face}
M.~G{\"u}nther, L.~El~Shafey, S.~Marcel, Face recognition in challenging
  environments: An experimental and reproducible research survey, in: Face
  Recognition Across the Imaging Spectrum, Springer, 2016, pp. 247--280.

\bibitem{huang2017beyond}
R.~Huang, S.~Zhang, T.~Li, R.~He, Beyond face rotation: Global and local
  perception gan for photorealistic and identity preserving frontal view
  synthesis, ICCV, 2017.

\end{thebibliography}
}

\end{document}